\documentclass{article}
\usepackage{ijcai18}

\usepackage{times}
\usepackage{xcolor}
\usepackage{soul}
\usepackage[utf8]{inputenc}
\usepackage[small]{caption}

\usepackage{graphicx}
\usepackage{color}
\usepackage{amsmath}
\usepackage{amsfonts} 
\usepackage{mathptmx}
\usepackage{bm}
\usepackage{url}
\usepackage{algorithm}
\usepackage{algorithmic}
\usepackage{multicol,lipsum}
\usepackage{cuted}
\usepackage{flushend}
\newcommand{\argmax}{\mathop{\rm arg~max}\limits}
\newcommand{\argmin}{\mathop{\rm arg~min}\limits}
\usepackage{etoolbox}
\patchcmd{\thebibliography}
  {\settowidth}
  {\setlength{\itemsep}{0pt plus 0.0pt}\settowidth}
  {}{}
\apptocmd{\thebibliography}
  {\small}
  {}{}
\title{Refining Manually-Designed Symbol Grounding and High-Level Planning \\ by Policy Gradients}

\author{
Takuya Hiraoka$^1$$^2$, 
Takashi Onishi$^1$$^2$,  
Takahisa Imagawa$^2$,
Yoshimasa Tsuruoka$^2$$^3$,  
\\ 
$^1$ NEC Central Research Laboratories \\
$^2$ National Institute of Advanced Industrial Science and Technology \\
$^3$ The University of Tokyo \\
t-hiraoka@ce.jp.nec.com,
t-onishi@bq.jp.nec.com, 
tsuruoka@logos.t.u-tokyo.ac.jp
}

\begin{document}

\maketitle

\begin{abstract}
Hierarchical planners that produce interpretable and appropriate plans are desired, especially in its application to supporting human decision making. 
In the typical development of the hierarchical planners, higher-level planners and symbol grounding functions are manually created, and this manual creation requires much human effort.
In this paper, we propose a framework that can automatically refine symbol grounding functions and a high-level planner to reduce human effort for designing these modules. 
In our framework, symbol grounding and high-level planning, which are based on manually-designed knowledge bases, are modeled with semi-Markov decision processes. 
A policy gradient method is then applied to refine the modules, in which two terms for updating the modules are considered. 
The first term, called a reinforcement term, contributes to updating the modules to improve the overall performance of a hierarchical planner to produce appropriate plans. 
The second term, called a penalty term, contributes to keeping refined modules consistent with the manually-designed original modules. Namely, it keeps the planner, which uses the refined modules, producing interpretable plans. 
We perform preliminary experiments to solve the Mountain car problem, and its results show that a manually-designed high-level planner and symbol grounding function were successfully refined by our framework. 
\end{abstract}

\section{Introduction}\label{sec:intro}
Hierarchical planners have been widely researched in artificial intelligence communities. 
One of the main reasons for that is that the hierarchical planers can divide complex planning problems, which flat planners cannot solve, into a series of more simple sub-problems, by using high-level knowledges about the planning problem (e.g., ~\cite{nilsson1984shakey,choi2009combining,kaelbling2011hierarchical}). 

A hierarchical planner is composed of multiple planner layers that are typically divided into two types: high-level and low-level. 
A low-level planner performs micro-level planning, and it deals with raw information about an environment. 
In contrast, a high-level planner performs macro-level planning, and it deals with more abstract symbolic information. 
The raw and abstract symbolic information are mapped to each other by {\it symbol grounding} functions. 
Imagine that a hierarchical planner is used for controlling a humanoid robot to put a lemon on a board. 
Here, the high-level planner makes a plan such a ``Pick a lemon up, and then put it on a board." 
The low-level planner makes a plan for controlling the robot's motors according to sensor inputs, to achieve sub-goals given by the high-level planner (e.g., ``Pick a lemon up"). 
As the low-level planner cannot understand what ``Pick a lemon up" means, the symbol ground function converts it into actual values, in the environment, which the low-level planner can understand. 

Hierarchical planners are often used for supporting human decision making (e.g., in supply chain~\cite{ozdamar1998hierarchical} or clinical operations~\cite{fdez2011supporting}). 
In such cases, people make decisions on the basis of a plan, and thus it is necessary that 1) they understand the plan (especially one of a high-level planner) and 2) they can reach satisfying outcomes by following the plan (i.e., the hierarchical planner gives appropriate plans). 

In many previous studies on hierarchical planners, symbol grounding functions and high-level planners were designed manually~\cite{nilsson1984shakey,malcolm1990symbol,cambon2009hybrid,choi2009combining,dornhege2009integrating,wolfe2010combined,kaelbling2011hierarchical}. 
Although this makes it possible for people to understand the plans easily, much human effort is needed to carefully design a hierarchical planner that provides appropriate plans. 

Konidaris et al.~\shortcite{konidaris2014constructing,konidaris2015symbol,konidaris2016constructing} have proposed frameworks for automatically constructing symbol grounding functions and high-level planners, but they require a human to carefully analyze them to understand the plans. 
These constructed modules are often complicated and, in such cases, analysis becomes a burden. 

In this paper, we propose a framework that automatically refines manually-designed symbol grounding functions and high-level planners, with a policy gradient method. 
Our framework differs from frameworks proposed in the aforementioned previous studies on the basis of the following points: 
\begin{itemize}
\setlength{\parskip}{0cm}
\setlength{\itemsep}{0cm}
\item Unlike the hierarchical planners based solely on manually-designed symbol grounding functions and high-level planners~\cite{nilsson1984shakey,malcolm1990symbol,cambon2009hybrid,choi2009combining,dornhege2009integrating,wolfe2010combined,kaelbling2011hierarchical}, our framework refines these modules without human intervention. 
This automated refinement reduces the design workload for the modules. 

\item Unlike the frameworks that automatically construct symbol grounding functions and high-level planners~\cite{konidaris2014constructing,konidaris2015symbol,konidaris2016constructing}, our framework refines these while attempting to keep the resulting symbol grounding consistent with prior knowledge of the definition of the symbols as much as possible (see Section \ref{sec:refinment}). 
Therefore, a person can understand the plan that high-level planners output without careful analysis of the refined modules. 
\end{itemize}
In this paper, we first explain our hierarchical planner (including the high-level planner and symbol grounding functions), and how these are designed (Section \ref{sec:hplanner}). 
Then, we introduce the framework designed to refine them (Section \ref{sec:refinment}). 
Finally, we experimentally demonstrate the effectiveness of our framework (Section \ref{sec:experiment}).

\section{Preliminaries}\label{sec:prel}
Our framework, introduced in Section \ref{sec:refinment}, is based on semi-Markov decision processes (SMDPs) and policy gradient methods. 

\subsection{Semi-Markov Decision Processes}\label{sec:smdp}
SMDPs are a framework for modeling a decision problem in an environment where a sojourn time in each state is a random variable, and it is defined as a tuple $\langle S, O, R, P, \gamma \rangle$. 
$S \subseteq \mathbb{R}^n$ is the $n$-dimensional continuous state space; $O(s)$ is a function that returns a finite set of options~\cite{sutton1999between} available in the environment's state $s \in S$; $R(s', t|s, o)$ is the reward received when option $o \in O(s)$ is executed at $s$; arriving in state $s' \in S$ after $t$ time steps; $P(s', t | s, o)$ is a probability of $s' \in S$, and $t$ after executing $o$ in $s$; and $\gamma \in [0, 1]$ is a discount factor. 

Given SMDPs, our interest is to find an optimal policy over options $\pi^*(o|s)$: 
\begin{eqnarray}
\pi^* &=& \argmax_{\pi} V_{\pi}(s_0), \label{eq:pistar} \\
V_{\pi}(s_0) &=& E_{\pi}\begin{bmatrix}R(s_1 , t_0 |s_0, o_0) + \gamma^tR(s_2 , t_1 |s_1, o_1) + ... \end{bmatrix} \label{eq:value}, 
\end{eqnarray}
where $(s_0, o_0, t_0, s_1)$ and $(s_1, o_1, t_1, s_2)$ are transitions of a state, an option, time steps elapsed while executing the option, and the arriving state after executing the option. 

\subsection{Policy Gradient}\label{sec:pg}
To find $\pi^*$, we use a policy gradient method~\cite{sutton2000policy}. 
In a policy gradient method, a policy $\pi_{\theta}(o|s)$ parameterized by $\theta$ is introduced to approximate $\pi^*$, and the approximation is performed by updating $\theta$ with a gradient. 
Although there are many policy gradient implementations (e.g., \cite{kakade2002natural,silver2014deterministic,schulman2015trust}), we use REINFORCE~\cite{williams1992simple}. 
In REINFORCE, $\theta$ is updated as follows: 
\begin{eqnarray}
\theta &\leftarrow& \theta + \alpha \nabla_{\theta} \log \pi_{\theta}(\tilde{s_0}, \tilde{o_0}) V_{\pi_{\theta}}(\tilde{s_0}), \label{eq:up} \\
V_{\pi_{\theta}}(\tilde{s_0}) &=& R(\tilde{s_1}, \tilde{t_0}| \tilde{s_0}, \tilde{o_0}) + \gamma^{\tilde{t_0}} R(\tilde{s_2}, \tilde{t_1}| \tilde{s_1}, \tilde{o_1}) + ... \label{eq:valapp}, 
\end{eqnarray}
where $\alpha$ is a learning rate and $(\tilde{s_0}, \tilde{o_0}, \tilde{t_0}, \tilde{s_1})$ and $(\tilde{s_1}, \tilde{o_1}, \tilde{t_1}, \tilde{s_2})$ are transitions of state, the executing option, elapsed time steps, and arriving state, which are sampled on the basis of $\pi_{\theta}$ in a time horizon. 
Other variables and functions are the same as those introduced in Section \ref{sec:smdp}. 
We decided to use REINFORCE for our work because it has successfully worked in recent work~\cite{silver2016mastering,das2017learning}.

\section{Hierarchical Planner with Symbol Grounding Functions}\label{sec:hplanner}
In this section, we first describe the outline of a hierarchical planner (including the high-level planner) with symbol grounding functions, which are manually designed. 
We then provide concrete examples of them. 
The high-level planner and symbol grounding functions described here are refined by the framework, which is proposed in Section \ref{sec:refinment}. 

\begin{figure*}[t]
\begin{center}
\includegraphics[width=1.0\hsize]{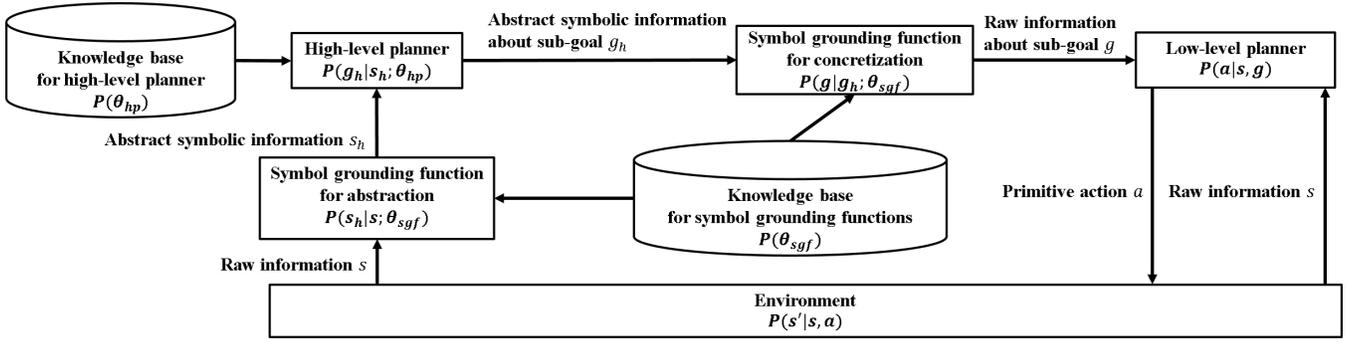}
\vspace{-8pt}
\caption{Outline of hierarchical planner with grounding functions.}
\label{fig:hplanner}
\end{center}
\vspace{-10pt}
\end{figure*}

The hierarchical planner (Figure \ref{fig:hplanner}) is composed of two symbol grounding functions (one for abstraction and the other for concretization), a high-level planner, a low-level planner, and two knowledge bases (one each for the high-level and low-level planners). 
These modules work as follows: 
\begin{description}
\setlength{\parskip}{0cm}
\setlength{\itemsep}{0cm}
\item[Step 1]: 
The symbol grounding function for abstraction receives raw information, abstracts it to a symbolic information on the basis of its knowledge base, and then outputs the symbolic information. 
\item[Step 2]: 
The high-level planner receives the abstract symbolic information, makes a plan using its knowledge base, and then outputs abstract symbolic information as a sub-goal, which indicates the next abstract state to be achieved. 
\item[Step 3]: 
The symbol grounding function for concretization receives the abstract symbolic information, concretizes it to raw information about a sub-goal, which specifies an actual state to be achieved, then outputs the raw information on the sub-goal. 
This module performs the concretization on the basis of its the knowledge base. 
\item[Step 4]: 
The low-level planner receives the raw information about a sub-goal  and then interacts with the environment to achieve the given sub-goal. 
In the interaction, the low-level planner outputs primitive actions in accordance with the raw information given by the environment. 
The interaction continues until the low-level planner achieves the given sub-goal, or until the total number of elapsed time steps reaches a given threshold. 
\item[Step 5]: 
If the raw information from the environment is not a goal or terminal state, return to the Step 1. 
\end{description}

The knowledge bases for symbol grounding functions and the high-level planners are designed manually. 
%
\begin{description}
\setlength{\parskip}{0cm}
\setlength{\itemsep}{0cm}
\item[Knowledge base for high-level planners] is described as a simple planning domain definition language (PDDL)~\cite{mcdermott1998pddl}. 
In a PDDL, objects, predicates, goals, and operators are manually specified. 
The objects and predicates are for building logical formulae, which specify the possible states in the planning domain. 
The operators are represented as a pair of preconditions and effects. 
The preconditions represent the states required for applying the operator, and the effects represent the arriving states after applying the operators. 
We use PDDLs in this work because they are widely used for describing a knowledge base for symbolic planners. 

\item[Knowledge base for symbol grounding functions] is described as a list of maps between abstract symbolic information and corresponding raw information. 
In this paper, to simplify the problem, we assume that each item of abstract symbolic information is mapped into one interval of raw information. 
Despite its simplicity, it is useful for representing, for example, typical spatial information. 
\end{description}

\begin{figure}[t]
\begin{center}
\includegraphics[width=0.5\hsize]{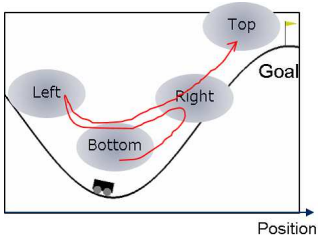}
\vspace{-8pt}
\caption{Mountain car with abstract symbols.}
\label{fig:mountaincar}
\end{center}
\vspace{-10pt}
\end{figure}

Here, we describe the knowledge bases and how the hierarchical planner works to solve the mountain car problem~\cite{moore1990efficient} (Figure \ref{fig:mountaincar}). 
In this problem, a car is placed within a deep valley, and its goal is to drive out by going up the right side hill. 
However, as the car's engine is not strong enough, it needs to first drive back and forth between the two hills to generate momentum. 
In this problem, the hierarchical planner receives raw information (the position and velocity of the car) from the environment and is required to make a plan to move it to the goal (the top of the right side hill). 

An example of knowledge for the high-level planner is shown in Table \ref{tab:kbhp}. 
In this example, objects are composed of only a ``Car." Predicates are composed of four instances (``Bottom\_of\_hills($x$), On\_right\_side\_hill($x$), On\_left\_side\_hill($x$), and At\_top\_of\_right\_side\_hill($x$)").
For example, ``On\_right\_side\_hill(Car)" means that the car is on the right side hill. 
Operators are composed of three types that refer to a transition of the objects on the hills. 
For example, ``Opr.1" refers to the transition that object $x$ has moved from the bottom of the hills to the right side hill. 
\begin{table}[t]
\centering
\caption{Example knowledge for high-level planners. Upper part describes examples of objects, predicates, and goals. Lower part describes examples of operators. }
\vspace{-8pt}
\label{tab:kbhp}
\scalebox{0.9}{\begin{tabular}{l|l}\hline
 Objects & $x = \{$Car$\}$ \\\hline
 Predicates & Bottom\_of\_hills($x$), On\_right\_side\_hill($x$), \\
 & On\_left\_side\_hill($x$), At\_top\_of\_right\_side\_hill($x$) \\\hline
Goals & At\_top\_of\_right\_side\_hill(Car) \\\hline
\end{tabular}} 
\\\vspace{7pt}
\scalebox{0.9}{\begin{tabular}{l|l|l}\hline
 Operators & Preconditions & Effects \\\hline\hline
 Opr.1 & Bottom\_of\_hills($x$) & On\_right\_side\_hill($x$) \\\hline
 Opr.2 & On\_right\_side\_hill($x$) & On\_left\_side\_hill($x$) \\\hline
 Opr.3 & On\_left\_side\_hill($x$) & At\_top\_of\_right\_side\_hill($x$) \\\hline
\end{tabular}}
\end{table}

An example of the knowledge for symbol grounding functions is shown in Table \ref{tab:kbsgf}. 
This example shows mappings between abstract symbolic information (the location of the car), and corresponding intervals of raw information (the actual value of the car's position). 
For example, ``Bottom\_of\_hills(Car)" is mapped to the position of the car is in the interval [-0.6, -0.4]. 
\begin{table}[t]
\centering
\caption{Example knowledge for symbol grounding functions. }
\vspace{-8pt}
\label{tab:kbsgf}
\scalebox{0.9}{\begin{tabular}{l|l}\hline
 Abstract symbolic informations & Interval of raw information \\\hline\hline
 Bottom\_of\_hills(Car) & position $\in [-0.6, -0.4]$ \\\hline
 On\_right\_side\_hill(Car) & position $\in [-0.2, 0.4]$ \\\hline
 On\_left\_side\_hill(Car) & position $\in [-1.2, -0.8]$ \\\hline
 At\_top\_of\_right\_side\_hill(Car) & position $\in [0.6, 0.8]$ \\\hline
\end{tabular}} 
\end{table}

Given the knowledge described in Tables \ref{tab:kbhp} and \ref{tab:kbsgf}, an example of how the hierarchical planner works is shown as follows: 
\begin{description}
\setlength{\parskip}{0cm}
\setlength{\itemsep}{0cm}
\item[Example of Step 1:] 
The symbol grounding function for abstraction receives raw information (position=-0.5 and velocity=0). 
The position is in the interval [-0.6, -0.4], which corresponds to the ``Bottom\_of\_hills(Car)" in Table \ref{tab:kbsgf}. 
Therefore, the symbol grounding function outputs ``Bottom\_of\_hills(Car)." 

\item[Example of Step 2:] 
The high-level planner receives ``Bottom\_of\_hills(Car)," and makes a plan to achieve the goal (``At\_top\_of\_right\_side\_hill(Car)"). 
By using the knowledge in Table \ref{tab:kbhp}, the high-level planner makes the plan [Bottom\_of\_hills(Car) $\rightarrow$ On\_right\_side\_hill(Car) $\rightarrow$ On\_left\_side\_hill(Car) $\rightarrow$ At\_top\_of\_right\_side\_hill(Car)], which means ``Starting at the bottom of the hills, visit, in order, the right side hill, the left side hill, and the top of the right side hill." 
After following the plan, the high-level planner outputs ``On\_right\_side\_hill(Car)." 

\item[Example of Step 3:] 
The symbol grounding function receives ``On\_right\_side\_hill(Car)," and concretizes it to raw information about sub-goal (position= 0.1, velocity=*). 
Here, the position in the raw information is determined as the mean of the corresponding interval $[-0.2, 0.4]$ in Table \ref{tab:kbsgf}. 
In addition, the mask (represented by ``*") is putted to filter out factors in raw information, which is irrelevant in the sub-goal (i.e., velocity in this example). 

\item[Example of Step 4:] 
The low-level planner receives position= 0.1 and the mask. 
To move the car to the given sub-goal (position=0.1), the low-level planner makes a plan to accelerate the car. 
This planning is performed by model predictive control~\cite{camacho2013model}. 
The low-level planner terminates itself when the car arrives at the given sub-goal (position=0.1), or when it takes a primitive action 20 times. 
\end{description}

\section{Framework for Refining Grounding Function and High Level Planner}\label{sec:refinment}
In this section, we propose a framework for refining the symbol grounding functions and the high-level planner introduced in the previous section. 
In our framework, symbol grounding and high-level planning, which are based on manually-designed knowledge bases, are modeled with SMDPs.  
Refinement of the symbol grounding functions and the high-level planner is achieved by applying policy gradients to the model. 
First, we introduce an abstract model and then provide an example of its implementation in the mountain car problem. 
Finally, we explain how the policy gradient method is applied to the model. 

\subsection{Modeling Symbol Grounding and High-Level Planning with SMDPs}
We model symbol grounding and high-level planning, which are based on manually-designed knowledge bases, with SMDPs. 
The symbol grounding functions and the high-level planner are modeled as components of the parameterized policy. 
In addition, the knowledge bases are modeled as priors for the policy's parameters. 

We first assume that information and modules, which appear in hierarchical planning, are represented as random variables and probability functions, respectively (Figure \ref{fig:hplanner}). 
Suppose that $S_h$ is a set of all possible symbols the symbol grounding functions and the high-level planner deal with, raw information is represented as an $n$-dimensional vector, and $A \subset \mathbb{R}^m$ is a set of all possible primitive actions. 
We denote raw information by $s, s' \in \mathbb{R}^n $ \footnote{The denotation is the same as that of the state described in Section \ref{sec:smdp} because the raw information is modeled as the state. }, abstract symbol information by $s_h \in S_h$, abstract symbolic information about a sub-goal by $g_h \in S_h$, raw information about a sub-goal $g \in \mathbb{R}^n$, and a primitive action by $a \in A$. 
In addition, we denote the symbol grounding function for abstraction by $P(s_h | s; \theta_{sgf})$, the symbol grounding function for concretization by $P(g | g_h ; \theta_{sgf})$, the high-level planner by $P(g_h | s_h ; \theta_{hp})$, the low-level planner by $P(a | s, g)$, the environment by $P(s' | s, a)$, the knowledge base for high-level planners by $P(\theta_{hp})$, and the knowledge base for the high-level planner by $P(\theta_{sgf})$. 
Here, $\theta_{sgf}$ and $\theta_{hp}$ are the parameters for the symbol grounding functions and the high-level planner, respectively. 

\begin{figure}[t]
\begin{center}
\includegraphics[width=1.0\hsize]{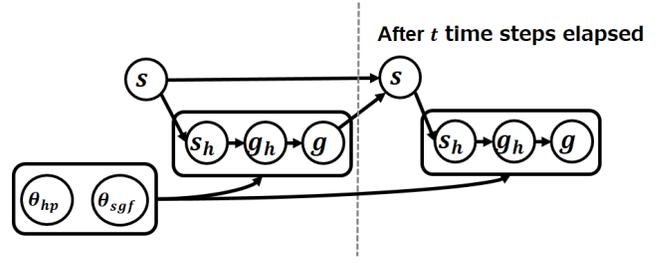}
\vspace{-13pt}
\caption{SMDPs for our framework.}
\label{fig:smdps}
\end{center}
\vspace{-10pt}
\end{figure}

High-level planning and symbol grounding based on the knowledge base are modeled as SMDPs (Figure \ref{fig:smdps}). 
In this model, the components of SMDPs (i.e., an option, a state, a reward, and a transition probability) are implemented as follows: 
\begin{description}
\setlength{\parskip}{0cm}
\setlength{\itemsep}{0cm}
\item[Option $o$:] is implemented as a tuple $\langle s_h, g_h, g \rangle$ of abstract symbolic information $s_h$, abstract symbolic information about a sub-goal $g_h$, and raw information about a sub-goal $g$. 
\item[State $s$:] is implemented as raw information. 
\item[Reward $R(s', t | s, o)$:] is the cumulative reward given by the environment $P(s' | s, a)$, while the low-level planner $P(a|s, g)$ is interacting with $P(s' | s, a)$. 
\item[Transition probability $P(s', t | s, o)$:] is implemented as a function, which represents the state transition proceeded by the interaction between the low-level planner $P(a|s, g)$ and the environment $P(s' | s, a)$. Note that although the transition probability receives option $\langle s_h, g_h, g \rangle$, only $g$ is used in the transition probability. 
\end{description}

In this model, the parameterized policy $\pi_{\theta}$ is implemented to control abstraction of raw information, high-level planning, and concretization of abstract symbolic information, in accordance with the knowledge bases. 
Formally, $\pi_{\theta}$ is implemented as follows: 
\begin{eqnarray}
\pi_{\theta}(g, g_h, s_h | s) &=& P(g, g_h, s_h, \theta_{sgf}, \theta_{hp} | s) \nonumber \\
 &=& P(g | g_h ; \theta_{sgf}) P(g_h | s_h ; \theta_{hp}) P(s_h | s; \theta_{sgf}) \nonumber \\ 
 & & \cdot P(\theta_{sgf}) P(\theta_{hp}). \label{eq:ourpi}
\end{eqnarray}
The right term in the second line can be derived by decomposing the joint probability in the first line, in accordance with the probabilistic dependency shown in Figure \ref{fig:mountaincar}. 
Note that, in this equation, $\theta$ is represented as $\theta_{sgf} \vert \vert \theta_{hp}$, i.e., a concatenation of $\theta_{sgf}$ and $\theta_{hp}$. 
By using this representation for $\pi_{\theta}$, we can derive an update expression, which can refines $\theta_{sgf}$ and $\theta_{hp}$ keeping them consistent with $P(\theta_{sgf})$ and $P(\theta_{hp})$. 
See Section \ref{sec:refine} for details. 

$P(\theta_{sgf})$ and $P(\theta_{hp})$ are needed to reflect the manually-designed knowledge bases. 
To do so, first, $P(\theta_{sgf})$ and $P(\theta_{hp})$ are implemented as parametric distributions $P(\theta_{sgf} ; \theta^{'}_{sgf})$ and $P(\theta_{hp} ; \theta^{'}_{hp})$, respectively, and their hyper-parameters $\theta^{'}_{sgf}$ and $\theta^{'}_{hp}$ are determined to replicate manually-designed symbol grounding functions and high-level planners. 
More formally, we use $\theta^{'*}_{sgf}$ and $\theta^{'*}_{sgf}$ as the optimal parameters of $\theta^{'}_{sgf}$ and $\theta^{'}_{hp}$, respectively, acquired by the following equations: 
\begin{eqnarray}
\theta^{'*}_{sgf} &=& \argmin_{\theta^{'}_{sgf}} D_{sgf}\begin{pmatrix}P(g | g_h ; \theta_{sgf}) P(\theta_{sgf} ; \theta^{'}_{sgf})\end{pmatrix}, \label{eq:psgf} \\
\theta^{'*}_{hp} &=& \argmin_{\theta^{'}_{hp}} D_{hp}\begin{pmatrix}P(g_h | s_h ; \theta_{hp}) P(\theta_{hp} ; \theta^{'}_{hp})\end{pmatrix}, \label{eq:php}
\end{eqnarray}
where $D_{sgf}$ and $D_{hp}$ are a divergence (e.g., KL divergence) from the manually-designed symbol grounding function and high-level planner, respectively. $D_{sgf}$ and $D_{hp}$ are abstract criteria, and thus, there are many implementations of functionals ``$\argmin_{\theta^{'}_{sgf}} D_{sgf}\begin{pmatrix}\cdot\end{pmatrix}$" and ``$\argmin_{\theta^{'}_{hp}} D_{hp}\begin{pmatrix}\cdot\end{pmatrix}$." 

\subsection{An Example of Model Implementation to Solve the Mountain Car Problem}\label{sec:exmodel}
We introduced an abstract model for symbol grounding and high-level planning with knowledge bases in the previous section. In this section, we provide an example of an implementation of the model to solve the mountain car problem. 

First, $S_h$ and $A$ are implemented as follows: 
\begin{eqnarray}
S_h &:=& \begin{Bmatrix} \scalebox{0.8}{Bottom\_of\_hills(Car), On\_right\_side\_hill(Car),} \\
 \scalebox{0.8}{On\_left\_side\_hill(Car), At\_top\_of\_right\_side\_hill(Car)}\end{Bmatrix}, \\
A &:=& [-1.0, 1.0]. 
\end{eqnarray}
$S_h$ is implemented in accordance with the knowledge shown in Table \ref{tab:kbsgf}. 
$A$ is implemented in accordance with the definition of actions to solve the mountain car problem, and represented as a set of values for the acceleration of the car. 

Second, the probabilities of the modules in the hierarchical planner are implemented as follows: 
\begin{eqnarray}
P(s_h | s; \theta_{sgf}) &:=& \frac{N(s|\mu_{s_h}, \sigma_{s_h})}{\Sigma_{s_h' \in S_h} N(s|\mu_{s_h'}, \sigma_{s_h'})}, \label{eq:impsgf1} \\
P(g | g_h ; \theta_{sgf}) &:=& N(g|\mu_{g_h}, \sigma_{g_h}), \label{eq:impsgf2} \\
P(g_h | s_h ; \theta_{hp}) &:=& \frac{\exp(\phi(s_h)\bm{w_{g_h}^{T}})} {\Sigma_{g_h' \in S_h} \exp(\phi(s_h)\bm{w_{g_h'}^{T}})} \label{eq:imphp2}. 
\end{eqnarray}
$P(s_h | s; \theta_{sgf})$ is implemented as the normalized likelihood of a normal distribution (Eq. (\ref{eq:impsgf1})), and $P(g | g_h ; \theta_{sgf})$ is implemented as a normal distribution (Eq. (\ref{eq:impsgf2})). 
In Eq. (\ref{eq:impsgf1}) and Eq. (\ref{eq:impsgf2}), $N(s|\mu_{s_h}, \sigma_{s_h})$ represents a normal distribution for $s$, which is parameterized by mean $\mu_{s_h}$ and standard deviation $\sigma_{s_h}$, s.t, $\forall s_h \in S_h$. 
Note that $\mu_{s_h}$ and $\sigma_{s_h}$ are identical to $\mu_{g_h}$ and $\sigma_{g_h}$, respectively. 
$P(s_h | s; \theta_{sgf})$ is implemented as a softmax function (Eq. (\ref{eq:imphp2})). 
In Eq. (\ref{eq:imphp2}), $\phi(s_h)$ is a base function that returns a one-hot vector $\in \{0, 1\}^{|S_h|}$ in which only one element corresponding to the value of $s_h$ is set to a value of 1, and the other elements are set to a value of 0. 
$\bm{w_{g_h}} \in \mathbb{R}^{|S_h|}$ is a weight vector, s.t., $\forall g_h \in S_h$. 
In this implementation, $\theta_{sgf}$ is a vector composed of $\mu_{s_h}, \sigma_{s_h}$, s.t., $s_h \in S_h$, and $\theta_{hp}$ is a vector composed of the set $\bm{w_{g_h}}$, s.t., $g_h \in S_h$. $P(a | s, g)$ and $P(s' | s, a)$ are implemented as deterministic functions, which represent the simulator of environment and the model predictive controller. 

Third, the reward function $R(s', t | s, o)$ is implemented as follows: 
\begin{eqnarray}
R(s', t | s, o) &:=& \Sigma_{i=0}^{t} \gamma^{i} r(s_{i}, a_{i}), \label{eq:implrew} \\
r(s_i, a_i) &=& \begin{cases}
 100 & (\text{if car position in } s_i > 0.6) \\
 - a_i & (\text{otherwise})
 \end{cases} \label{eq:lowrew}, 
\end{eqnarray}
where $s_i$ and $a_i$ are a state and a primitive action sampled from the environment $i$ time steps later from the executing option $o$, respectively. 
Eq. (\ref{eq:lowrew}) represents ``low-level" reward $r(s_i)$, which is fed in accordance with $a_i$ and the car position included in $s_i$. 

Fourth, $P(\theta_{sgf} ; \theta^{'}_{sgf})$ and $P(\theta_{hp} ; \theta^{'}_{hp})$ are implemented as follows: 
\begin{eqnarray}
P(\theta_{sgf} ; \theta^{'}_{sgf}) &:=&\prod_{s_h \in S_h} N(\mu_{s_h} | \mu^{'}_{s_h}, 1) \cdot Uni(\sigma_{s_h}), \label{eq:priorsgf} \\
P(\theta_{hp} ; \theta^{'}_{hp}) &:=& \prod_{g_h \in S_h} \prod_{i = 0}^{|{s_h}|} N(w_{g_h, i} | \mu^{'}_{w_{g_h, i}}, 1). \label{eq:priorhp}
\end{eqnarray}
Eq. (\ref{eq:priorsgf}) represents a distribution for $\mu_{s_h}$ and $\sigma_{s_h}$. 
The component for $\mu_{s_h}$ is a normal distribution, which has mean $\mu^{'}_{s_h}$ and standard deviation 1, and the component for $\sigma_{s_h}$ is a uniform distribution $Uni(\sigma_{s_h})$. 
In addition, Eq. (\ref{eq:priorhp}) represents the normal distribution for $w_{g_h, i}$, which is the $i$-th element of $\bm{w_{g_h}}$. 
This distribution has mean $\mu^{'}_{w_{{g_h}, i}}$ and standard deviation 1. Note that, in this implementation, $\theta^{'}_{sgf}$ and $\theta^{'}_{hp}$ are $\mu^{'}_{s_h}$ and $\mu^{'}_{w_{g_h, i}}$, respectively. 

Finally, functionals in Eq. (\ref{eq:psgf}) and Eq. (\ref{eq:php}), are implemented as follows: 
\begin{description}
\setlength{\parskip}{0cm}
\setlength{\itemsep}{0cm}
\item[Implementation of $\argmin_{\theta^{'}_{sgf}} D_{sgf}\begin{pmatrix}\cdot\end{pmatrix}$ :] 
Using Eq. (\ref{eq:priorsgf}), $\mu_{s_h}'$ is set as the mean of the corresponding interval, which is defined in the knowledge base for grounding functions. 
For example, $\mu^{'}_{\text{Bottom\_of\_hills(car)}}$ is determined as $-0.5$, the mean of $[-0.6, -0.4]$ in Table \ref{tab:kbsgf}. 

\item[Implementation of $\argmin_{\theta^{'}_{hp}} D_{hp}\begin{pmatrix}\cdot\end{pmatrix}$:] 
Using Eq. (\ref{eq:php}), $\mu_{w_{g_h}}'$ is determined by Algorithm \ref{alg:initw}. 
The algorithm is outlined as follows: first initialize $\mu_{w_{g_h}}$ with $val_{nin}$ (line 1--3), and if the operator, in which $s_h$ refers to the preconditions and $s_h'$ refers to the effects, is contained in knowledge base $KB_{hp}$, the corresponding weight is initialized with $Val_{in}$ (line 4--11). 
$KB_{hp}$ is initialized in accordance with Table \ref{tab:kbhp} before it is passed to the algorithm. 
\end{description}
\begin{algorithm}[t]
\caption{Implementation of $\argmin_{\theta^{'}_{hp}} D_{hp}$}
\label{alg:initw}
\begin{algorithmic}[1]
\REQUIRE The following variables are given: \\ 
(1) Set of abstract symbolic information $S_h$. \\
(2) Set of operators $K_{hp}$ included in the knowledge base for the high-level planner. Each operator is represented as a tuple (precondition, effect). \\
(3) Set of hyper parameters $\mu_{w_{g_h}}'$ for all possible abstract symbolic information $s_h$. \\
(4) Weight value $val_{in}$ to be assigned to the weight of an operator, which is included in the knowledge base. \\
(5) Weight value $val_{nin}$ to be assigned to the weight of an operator, which is not included in the knowledge base. \\
(6) Index function $I(s_h)$ that maps $s_h$ to index $i \in \mathbb{I}$. $i$ used to access the element of $\mu_{w_{g_h}}'$. 
 \FOR{$s_h \in S_h$}
       \STATE{Initialize $\mu_{w_{g_h}}'$ with $val_{nin}$}
 \ENDFOR
 \FOR{$s_h \in S_h$}
         \STATE{$i \leftarrow 0$}
         \FOR{$s_h' \in S_h$}
                 \IF{$(s_h, s_h') \in K_{hp}$ }
                     \STATE{$\mu_{w_{g_h, I(s_h')}}' \leftarrow val_{in}$}
                 \ENDIF
         \ENDFOR
 \ENDFOR
\end{algorithmic}
\end{algorithm}

\subsection{Refining Symbol Grounding and High-Level Planning with Policy Gradients}\label{sec:refine}
Refining the high-level planner $P(g_h | s_h ; \theta_{hp})$ and symbol grounding functions ($P(s_h | s; \theta_{sgf})$ and $P(g | g_h ; \theta_{sgf})$) is achieved by a parameter update in Eq. (\ref{eq:update}). 
\begin{table*}[t]
\begin{eqnarray}
\theta_{sgf} \vert \vert \theta_{hp} \leftarrow \theta_{sgf} \vert \vert \theta_{hp} 
+ \alpha V_{\pi_{\theta_{sgf} \vert \vert \theta_{hp}}}(\tilde{s_0}) \{
\underbrace{\nabla_{\theta_{sgf} \vert \vert \theta_{hp}}\log P(\tilde{g_0} | \tilde{g_{h,0}} ; \theta_{sgf}) P(\tilde{g_{h,0}} | \tilde{s_{h,0}} ; \theta_{hp}) P(\tilde{s_{h,0}} | \tilde{s_0}; \theta_{sgf})}_{\text{reinforcement term}}
+ \underbrace{\nabla_{\theta_{sgf} \vert \vert \theta_{hp}}\log P(\theta_{sgf}) P(\theta_{hp})}_{\text{penalty term}}
\}
\label{eq:update}
\end{eqnarray}
\vspace{-15pt}
\end{table*}
This equation contains two unique terms: a \textbf{reinforcement term} and a \textbf{penalty term}. 
The reinforcement term contributes to updating the parameters to maximize the expected cumulative reward, as in standard reinforcement learning. 
The penalty term contributes to keeping the parameters consistent with the priors (i.e., manually-designed knowledge bases). 
This update is derived by substituting $\theta_{sgf} \vert\vert \theta_{hp}$ and Eq. (\ref{eq:ourpi}) for $\theta$ and Eq. (\ref{eq:up}), respectively. 
Using the example described in Section \ref{sec:exmodel}, $\mu_{s_h}$, $\sigma_{s_h}$ and $\bm{w_{g_h}}$ are updated in this equation. 
In this case, the penalty term prevents $\mu_{s_h}$ and $w_{g_h, i}$, for all $s_h$, $g_h$, and $i$, from moving far away from $\mu_{s_h}'$ and $\mu_{w_{g_h}, i}'$, respectively. 

\section{Experiments}\label{sec:experiment}
In this section, we perform an experimental evaluation to investigate whether the symbol grounding functions and high-level planner are refined successfully by using the framework we proposed in the previous section. 
In Section \ref{sec:evalgrounding}, we focus on the evaluation for refining the symbol grounding functions only. 
Then, in Section \ref{sec:evalgroundinghp}, we evaluate the effect of jointly refining symbol grounding functions and the high-level planner.

\subsection{Refinement of Symbol Grounding}\label{sec:evalgrounding}
We evaluate how the symbol grounding functions are refined by our framework to solve the mountain car problem. 
The experimental set up to implement the planner and our framework is the same as that in the example introduced in Section \ref{sec:hplanner} and Section \ref{sec:refinment}. 

For the evaluation, we prepared three types of method: 
\begin{description}
\setlength{\parskip}{0cm}
\setlength{\itemsep}{0cm}
\item[Baseline:] A hierarchical planner that uses the grounding functions and a high-level planner, which are manually designed. 
This planner is identical to that introduced in the example in Section \ref{sec:hplanner}. 
\item[NoPenalty:] The framework that refines the symbol grounding functions without the penalty term in Eq. (\ref{eq:update}). 
In this method, the high-level planner is the same as that in Baseline. 
\item[Proposed:] The framework that refines the symbol grounding functions with the penalty term. 
In this method, the high-level planner is the same as that in Baseline. 
\end{description}

These methods were evaluated on the basis of two criteria: an average cumulative reward over episodes, and a parameter divergence. 
The former is to evaluate if the hierarchical planner produces a more appropriate plan by refining its modules, and the latter is to evaluate the interpretability of the refined modules. 
The parameter divergence represents how much the policy's parameters ($\mu_{s_h}$)\footnote{We assume $\mu_{s_h}$ dominatingly determines the behaviors of symbol grounding functions. } refined by the framework differ from the initial parameters. 
In this paper, this divergence is measured by the Euclidean distance between the refined parameter ($\mu_{s_h}$) and its initial parameter ($\mu_{s_h}'$). 
Initial values for $\mu_{s_h}$ and $\sigma_{s_h}$ are given, shown as ``Init" in Table \ref{tab:rparamsgf}. 
$\mu_{s_h}$ is initialized with $\mu_{s_h}'$, which is determined on the basis of the implementation of the functional in Eq. (\ref{eq:psgf}) (see Section \ref{sec:exmodel}), and $\sigma_{s_h}$ is manually determined. 
We consider 50 episodes as one epoch and performed refinement over 2000 epochs. 

\begin{figure}[t]
\begin{minipage}{0.48\hsize}
\begin{center}
\includegraphics[width=1.0\hsize]{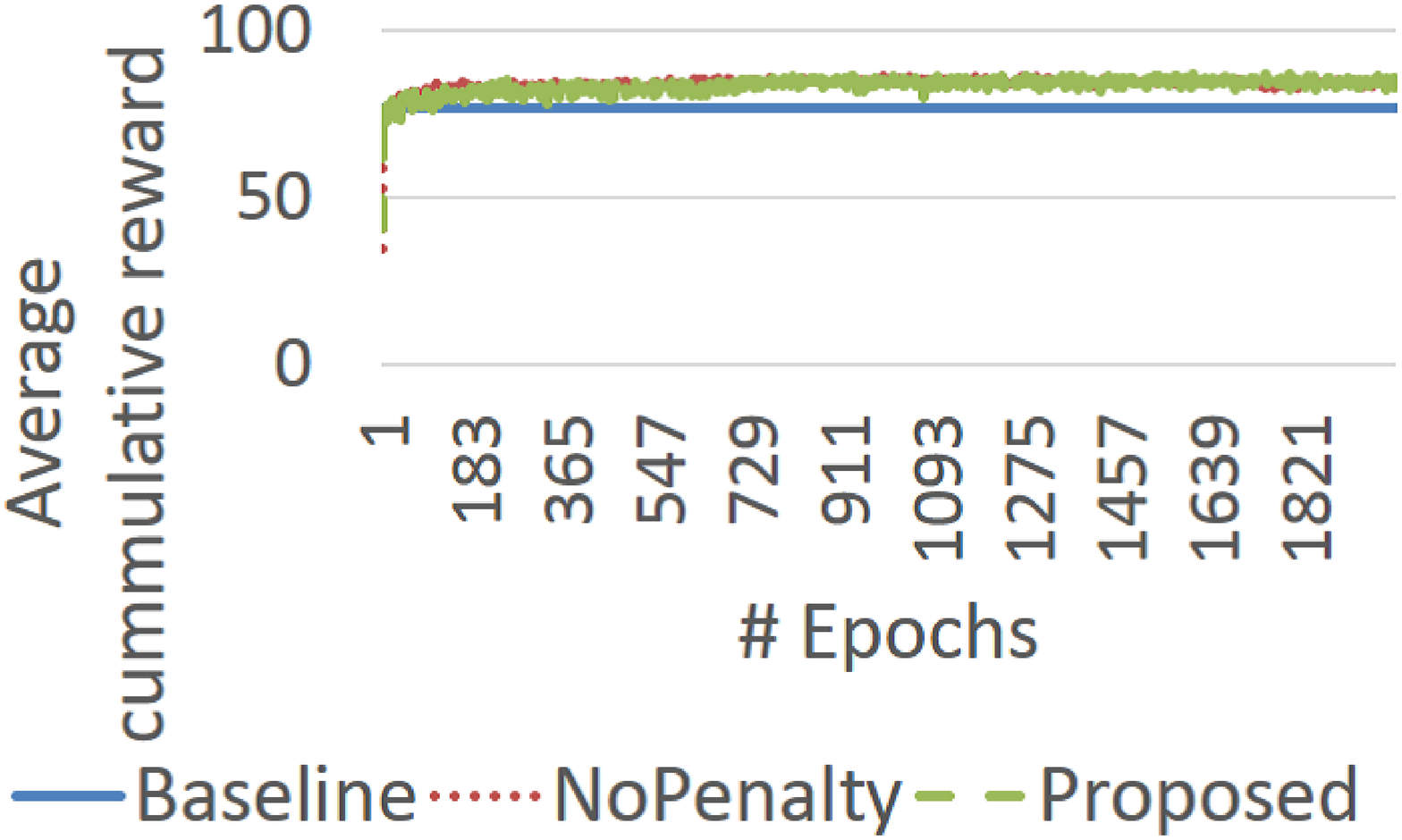}
\vspace{-18pt}
\caption{Learning curves for each methods. The vertical axis represents average cumulative rewards and the horizontal axis the horizontal axis represents epochs (50 episodes for each epoch). }
\label{fig:exp1avrrew}
\end{center}
\end{minipage}\hspace{9pt}\begin{minipage}{0.48\hsize}
\begin{center}
\includegraphics[width=1.0\hsize]{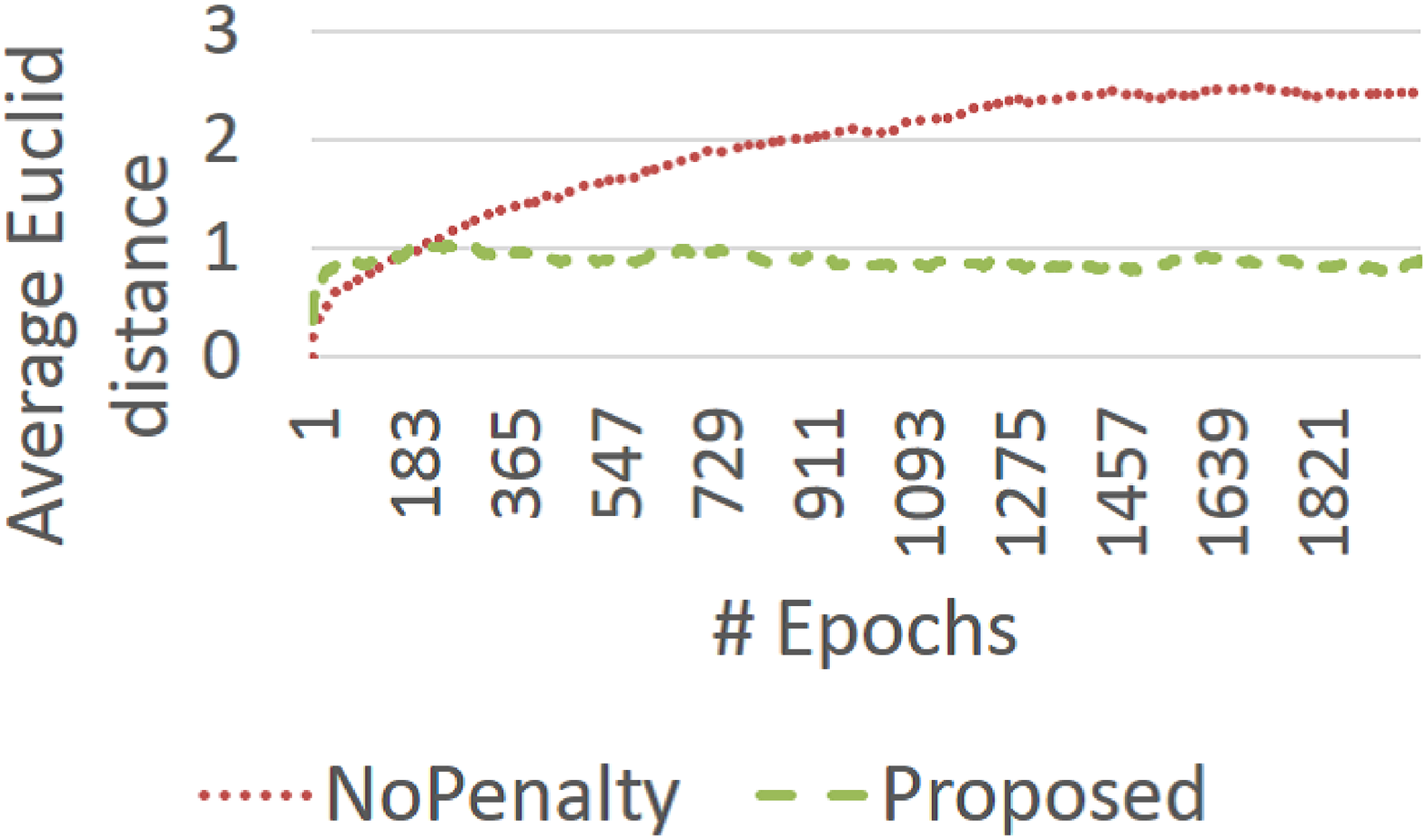}
\vspace{-18pt}
\caption{Parameter divergences. The vertical axis represents the average Euclidean distance and the horizontal axis represents learning epochs.}
\label{fig:exp1idst}
\end{center}
\end{minipage}
\end{figure}

The experimental results (shown in Figures \ref{fig:exp1avrrew} and \ref{fig:exp1idst}) show that 1) refining the grounding functions improves the performance (average cumulative reward) of hierarchical planners, and 2) considering the penalty term keeps the refined parameters within a certain distance from the initial parameters. 
Regarding 1), Figure \ref{fig:exp1avrrew} shows the methods in which the grounding functions are refined (NoPenalty and Proposed) outperform Baseline. This result indicates the refinement for grounding functions successfully improves its performance. 
Regarding 2), Figure \ref{fig:exp1idst} shows that the parameter in NoPenalty moves away from the original parameter in refining, while in Proposed, the parameter stays close to the original one. 

\begin{figure}[t]
\begin{minipage}{0.5\hsize}
\begin{center}
\includegraphics[width=1.0\hsize]{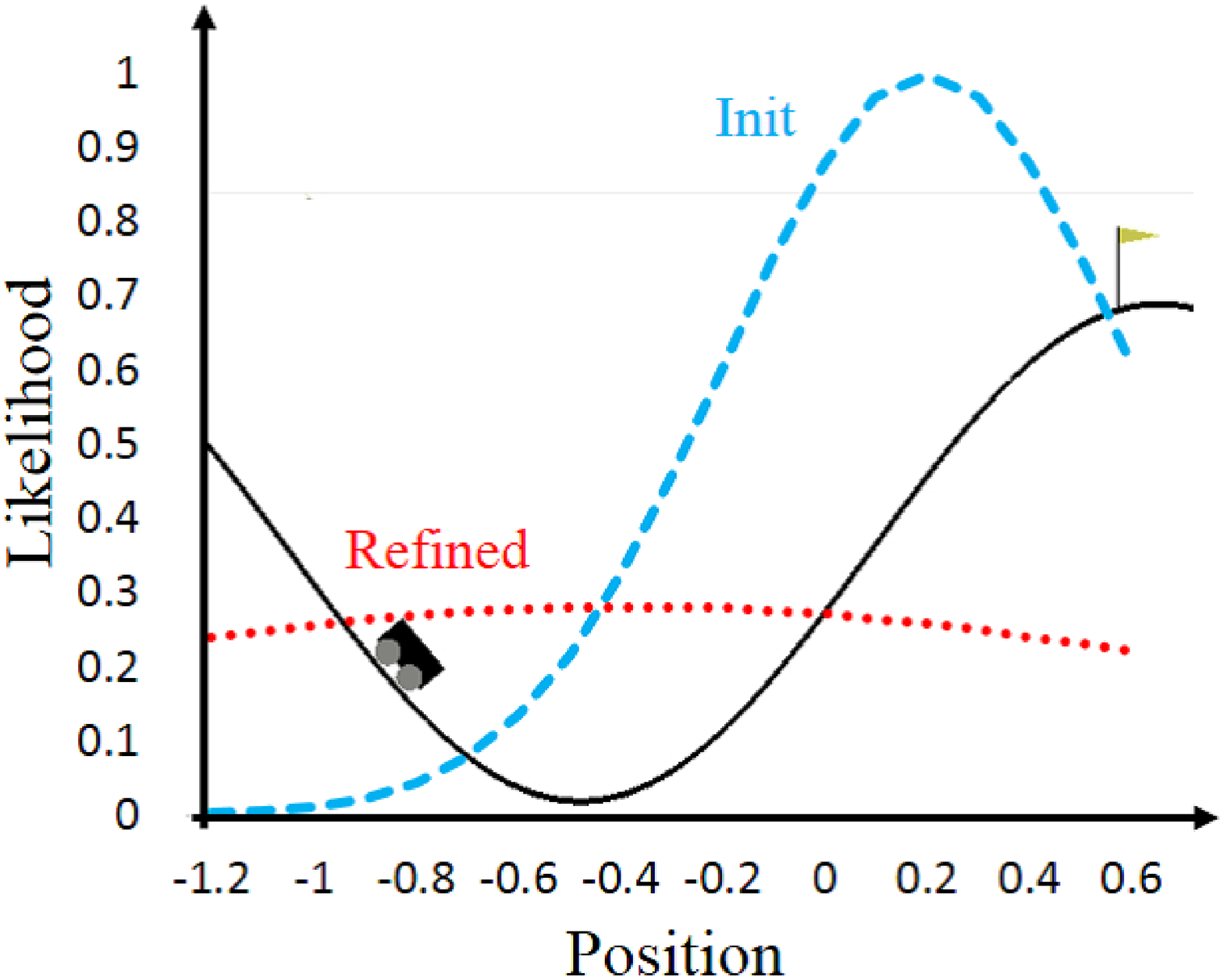}
\vspace{-18pt}
\caption{Example of refining result symbol grounding function for $On\_right\_side\_hill$. }
\label{fig:exp1weight}
\end{center}
\end{minipage}\begin{minipage}{0.5\hsize}
\begin{center} 
\includegraphics[width=1.0\hsize]{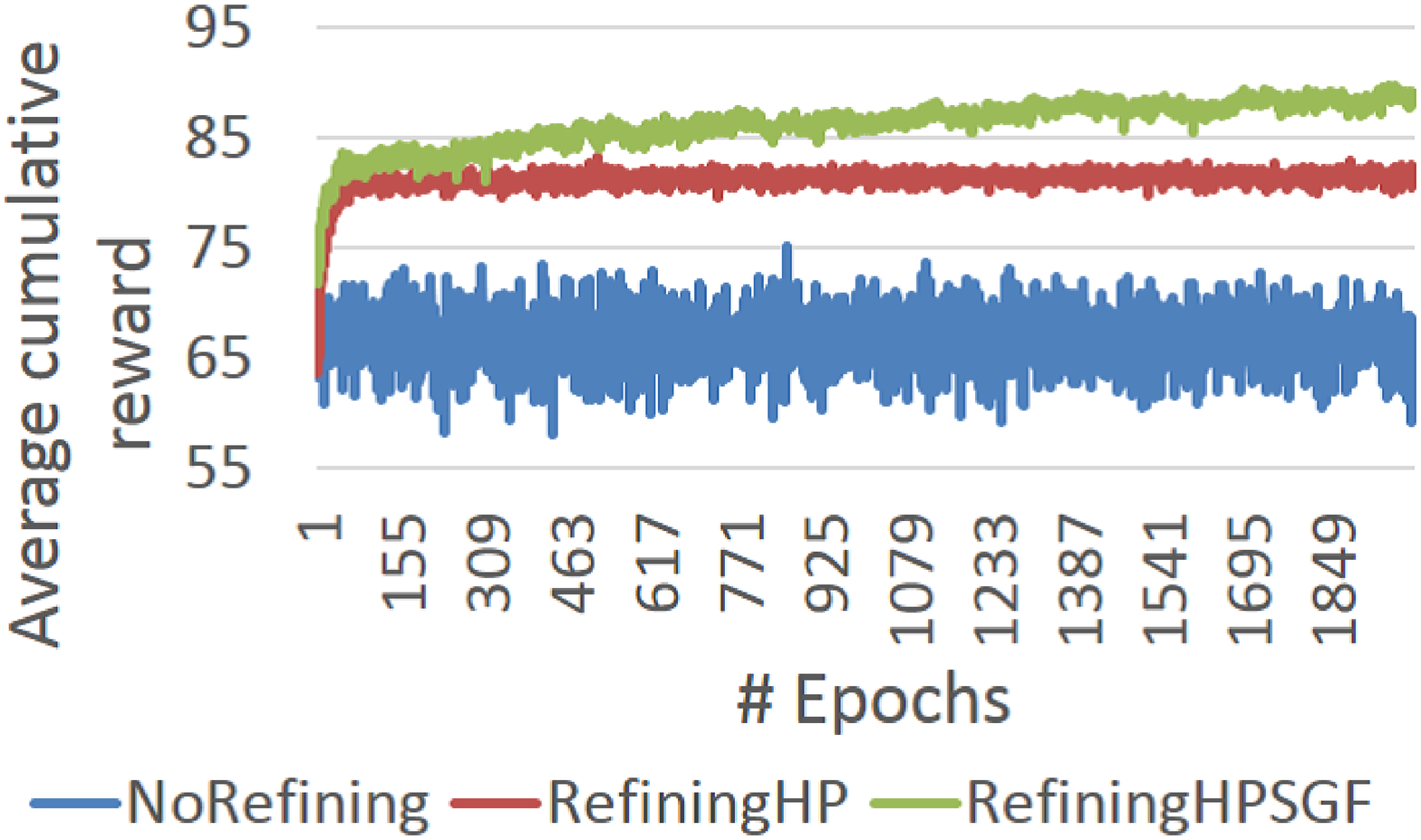}
\vspace{-18pt}
\caption{Learning curve.}
\label{fig:exp2avrrew}
\end{center}
\end{minipage}
\vspace{-10pt}
\end{figure}

An example of the refined parameter for the grounding functions for Proposed is shown in Table \ref{tab:rparamsgf}, which indicates that the parameter is updated to achieve high-performance planning while staying close to the original parameter. 
In this example, the mean and standard deviation of ``On\_right\_side\_hill(Car)" is changed significantly through refinement. 
The mean for grounding On\_right\_side\_hill(Car) is biased to a more negative position, and also flattened to make the car climb up the left side hill quickly (Figure \ref{fig:exp1weight}). 
This change makes the symbol grounding function more flattened and considers the center position as ``On\_right\_side\_hill(Car)." 
The main interpretation of this result is that the symbol grounding function was refined to reduce the redundancy in high-level planning. 
In the original symbol grounding functions, the center position is grounded to ``Bottom\_of\_hills (Car)," and the high-level planner makes a plan [Bottom\_of\_hills(Car) $\rightarrow$ On\_right\_side\_hill(Car) $\rightarrow$ On\_left\_side\_hill(Car) $\rightarrow$ At\_top\_of\_right\_side\_hill(Car)], which means ``Starting at the bottom of hills, visit, in order, the right side hill, the left side hill, and the top of the right side hill." 
However, this plan is redundant; the car does not need to visit the right side hill first. 
The refined symbol grounding function considers the center position as ``Right\_side\_hill(Car)," and thus the high-level planner produces the plan [Bottom\_of\_hills(Car) $\rightarrow$ On\_right\_side\_hill(Car) $\rightarrow$ On\_left\_side\_hill(Car) $\rightarrow$ At\_top\_of\_right\_side\_hill(Car)], in which the redundancy is removed. 
It should also be noted that the order of the refined means is intuitively correct. 
For example, the value of $\mu_{\text{On\_right\_side\_hill}}$ is higher than the value of $\mu_{\text{On\_left\_side\_hill}}$ (i.e., $\mu_{\text{On\_right\_side\_hill}}$ means the place on more right-side than $\mu_{\text{On\_left\_side\_hill}}$). 
It cannot be seen in the Baseline and NoPenalty cases. 
This result supports the fact that our framework refines the modules by maintaining their interpretability.
\begin{table}[t]
\centering
\caption{Refined parameters.}
\label{tab:rparamsgf}
\vspace{-8pt}
\scalebox{0.6}{\begin{tabular}{c||c|c|c|c}\hline
$\mu_{s_h}$ & $Bottom\_of\_hills$ & $At\_top\_of\_right\_side\_hill$ & $On\_right\_side\_hill$ & $On\_left\_side\_hill$ \\ \hline \hline
Init & -0.5 & 0.6 & 0.2 & -1.1 \\ \hline
Refined & -0.5 & 0.46 & \textbf{-0.39} & -1.1 \\ \hline
\end{tabular}}
\\ \vspace{5pt}
\scalebox{0.6}{\begin{tabular}{c||c|c|c|c}\hline
$\sigma_{s_h}$ & $Bottom\_of\_hills$ & $At\_top\_of\_right\_side\_hill$ & $On\_right\_side\_hill$ & $On\_left\_side\_hill$ \\ \hline \hline
Init & 0.4 & 0.1 & 0.4 & 0.3 \\ \hline
Refined & 0.4 & 0.12 & 1.42 & 0.11 \\ \hline
\end{tabular}}
\end{table}

\subsection{Joint Refinement of  Symbol Grounding and High-Level Planning}\label{sec:evalgroundinghp}
In this section, we refine both the symbol grounding functions and the high-level planner. 
The setup of the hierarchical planner and the problem are the same as those of the previous section, except for the knowledge base for the high-level planner. 
We removed ``Opr.2" (as shown in Table \ref{tab:kbhp}) and used this degraded version as the knowledge base for the experiment. 
This degradation makes a space for refining the knowledge base for the high-level planner. 
In addition, we put a small coefficient of the penalty term for $\bm{w_{g_h}}$, because we found that considering this term too much makes the refinement worse in a preliminary experiment. 
As long as the results of the symbol grounding functions are interpretable, the result of the high-level planner is interpretable as well. 
$\bm{w_{g_h}}$ is initialized with $w_{{g_h}, i}'$, which is determined by (i.e., Algorithm \ref{alg:initw}) where we set -1.3 as $val_{in}$, and -0.02 as $val_{nin}$. 
The resulting $w_{{g_h}, i}'$ is shown as ``Init" in Table \ref{tab:rparam}. 

We prepared three types of methods: 
\begin{description}
\setlength{\parskip}{0cm}
\setlength{\itemsep}{0cm}
\item[NoRefining:] 
A hierarchical planner with the degraded version of the knowledge base for high-level planner. 
The knowledge base for the symbol grounding function is the same to that shown in Table \ref{tab:kbsgf}. 
\item[RefiningHP:] 
The framework that refines the high-level planner only. 
In this method, symbol grounding functions are the same as those in NoRefining. 
\item[RefiningHPSGF:] 
The framework that refines both symbol grounding functions and the high-level planner. 
\end{description}

From the experimental result (Figure \ref{fig:exp2avrrew}), we can confirm that our framework successfully refines both symbol grounding functions and the high-level planner, from the viewpoint of performance. RefiningHP outperforms NoRefining, and RefiningHPSGF outperforms the other methods. 

Table \ref{tab:rparam} provides an example of how the high-level planner was refined. 
It indicates that the dropped knowledge (i.e., Opr. 2) was successfully acquired in refinement, and knowledge is discovered that makes high-level planning more efficient. 
Considering the form of Eq. (\ref{eq:imphp2}), the operator, which corresponds to the element of a weight with a higher value, contributes more to high-level planning. 
Therefore, these corresponding operators are worthwhile as knowledge for high-level planning. 
In Table \ref{tab:rparam}, the refined weight of the operator (preconditions=On\_right\_side\_hill, effects=On\_left\_side\_hill) is higher than those of other operators in which the precondition contains On\_right\_side\_hill. 
This operator was once initially removed and later acquired by the refinement. 
Similarly, the operator (preconditions=Bottom\_of\_hills, effects=On\_left\_side\_hill), which is not shown in Table \ref{tab:kbhp}, was newly acquired. 
\begin{table*}[t]
\centering
\caption{Example of high-level planner improvement. Refined weights $\bm{w_{g_h}}$ are shown for each precondition (column) and effect (row). Initial weights are shown in parentheses.}
\vspace{-9pt}
\label{tab:rparam}
\scalebox{0.95}{\begin{tabular}{c||c|c|c|c}\hline
Refined (Init) & Bottom\_of\_hills & At\_top\_of\_right\_side\_hill & On\_right\_side\_hill & On\_left\_side\_hill \\ \hline \hline
Bottom\_of\_hills & -5.88 (-1.3) & -6.34 (-1.3) & -3.15 (-1.3) & -6.65 (-1.3) \\ \hline
At\_top\_of\_right\_side\_hill & -9.04 (-1.3) & -9.75 (-1.3) & -4.76 (-1.3) & \textbf{2.5} (-0.02)  \\ \hline
On\_right\_side\_hill & -0.98 (-0.02) & \textbf{1} (-1.3) & -2.03 (-1.3) & -1.34 (-1.3) \\ \hline
On\_left\_side\_hill & \textbf{0.85} (-1.3) & -2.12 (-1.3) & \textbf{1.74} (-1.3) & -11.71 (-1.3) \\ \hline
\end{tabular}}
\vspace{-10pt}
\end{table*}

\section{Conclusion}
In this paper, we proposed a framework that refines manually-designed symbol grounding functions and a high-level planner. 
Our framework refines these modules with policy gradients. Unlike standard policy gradient implementations, our framework additionally considers the penalty term to keep parameters close to the prior parameter derived from manually-designed modules. 
Experimental results showed that our framework successfully refined the parameters for the modules; it improves the performance (cumulative reward) of the hierarchical planner, and keeps the parameters close to those derived from the manually-designed modules. 

One of the limitations of our framework is that it deals only with predefined symbols (such ``Bottom\_of\_hills"), and it does not discover new symbols. 
We plan to address this drawback in future work. 
We also plan to evaluate our framework in a more complex domain where primitive actions and states are high-dimensional, and the knowledge base is represented in a more complex description (e.g., precondition contains multiple states). 

\bibliographystyle{named}
\fontsize{1.0pt}{0pt}\selectfont
\bibliography{mybibAIST}
\end{document}